\documentclass[sigconf,natbib=true,anonymous=false]{acmart}
\usepackage{soul}
\usepackage{multirow}
\usepackage{subcaption}

\copyrightyear{2025}
\acmYear{2025}
\setcopyright{cc}
\setcctype{by}
\acmConference[WWW Companion '25]{Companion Proceedings of the ACM Web Conference 2025}{April 28-May 2, 2025}{Sydney, NSW, Australia}
\acmBooktitle{Companion Proceedings of the ACM Web Conference 2025 (WWW Companion '25), April 28-May 2, 2025, Sydney, NSW, Australia}
\acmDOI{XX.XXXX/XXXXXXX.XXXXXXX}
\acmISBN{979-8-4007-1331-6/2025/04}

\begin{document}

\title{The Impact of Persona-based Political Perspectives on Hateful Content Detection}

\author{Stefano Civelli}
\orcid{} 
\affiliation{%
  \institution{The University of Queensland}
  \city{Brisbane}
  \country{Australia}
}
\email{s.civelli@uq.edu.au}

\author{Pietro Bernardelle}
\orcid{0009-0003-3657-9229} 
\affiliation{%
  \institution{The University of Queensland}
  \city{Brisbane}
  \country{Australia}
}
\email{p.bernardelle@uq.edu.au}

\author{Gianluca Demartini}
\orcid{0000-0002-7311-3693} 
\affiliation{%
  \institution{The University of Queensland}
  \city{Brisbane}
  \country{Australia}
}
\email{demartini@acm.org}

\renewcommand{\shortauthors}{Stefano Civelli, Pietro Bernardelle, \& Gianluca Demartini}

\begin{abstract}
While pretraining language models with politically diverse content has been shown to improve downstream task fairness, such approaches require significant computational resources often inaccessible to many researchers and organizations. Recent work has established that persona-based prompting can introduce political diversity in model outputs without additional training. However, it remains unclear whether such prompting strategies can achieve results comparable to political pretraining for downstream tasks. We investigate this question using persona-based prompting strategies in multimodal hate-speech detection tasks, specifically focusing on hate speech in memes. Our analysis reveals that when mapping personas onto a political compass and measuring persona agreement, inherent political positioning has surprisingly little correlation with classification decisions. Notably, this lack of correlation persists even when personas are explicitly injected with stronger ideological descriptors. Our findings suggest that while LLMs can exhibit political biases in their responses to direct political questions, these biases may have less impact on practical classification tasks than previously assumed. This raises important questions about the necessity of computationally expensive political pretraining for achieving fair performance in downstream tasks.
\end{abstract}




\maketitle

\begin{figure}[t]
    \vspace{3mm}
    \centering
    \begin{subfigure}[b]{0.48\linewidth}
        \centering
        \includegraphics[width=\textwidth]{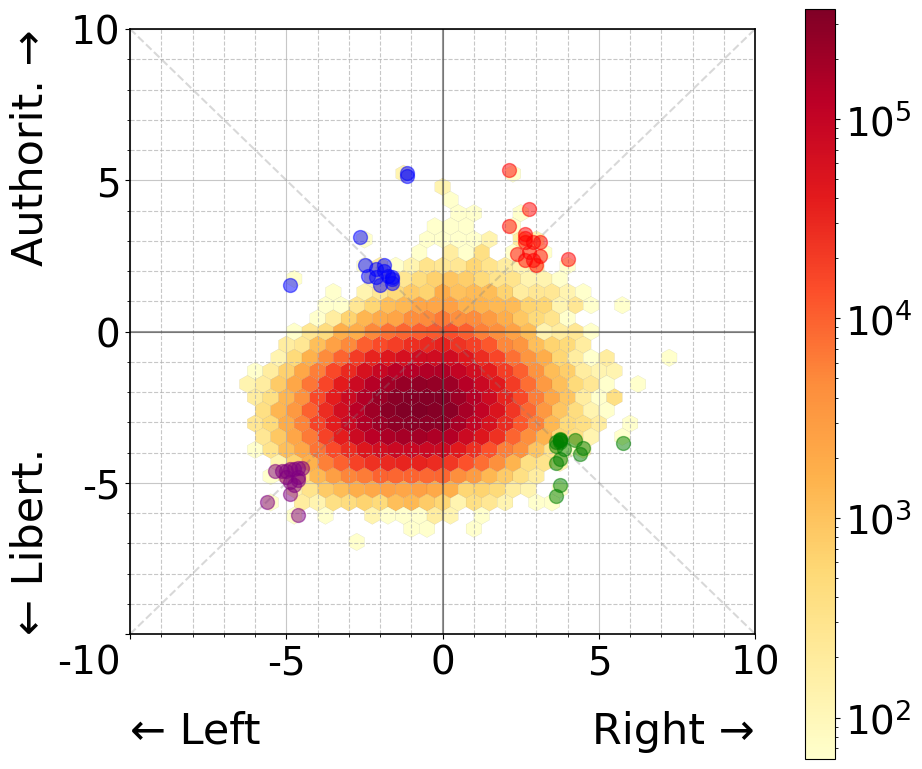}
        \caption{Distribution of four-corner extreme personas}
        \label{fig:diagonal}
    \end{subfigure}
    \hfill
    \begin{subfigure}[b]{0.48\linewidth}
        \centering
        \includegraphics[width=\textwidth]{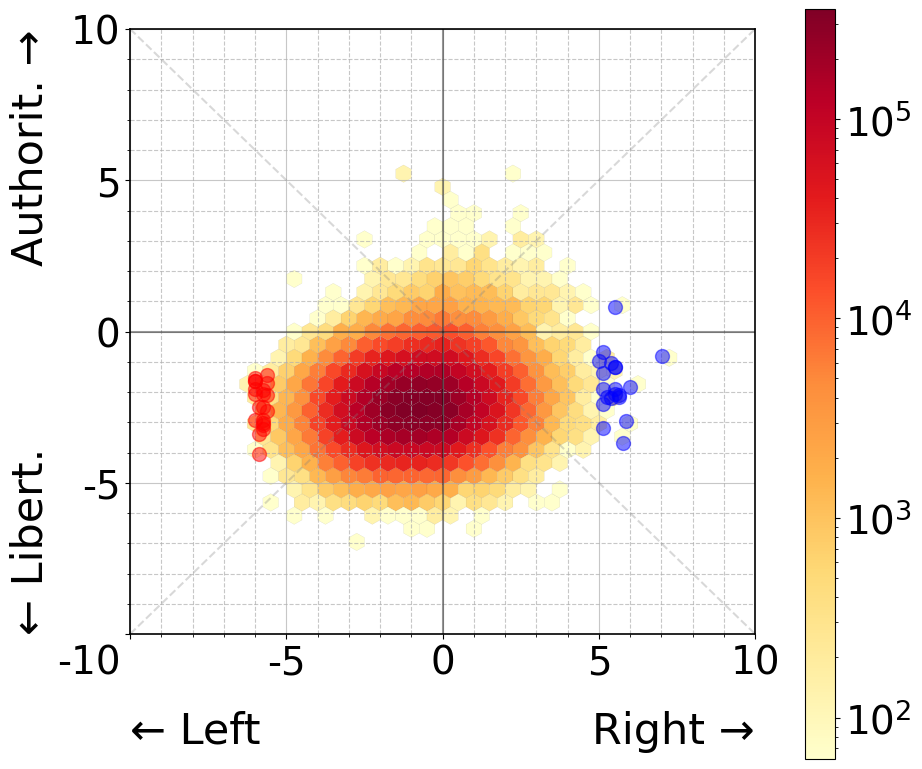}
        \caption{Distribution of left-right extreme personas}
        \label{fig:enter-label}
    \end{subfigure}
    \caption{Political compass distribution of PersonaHub personas when impersonated by IDEFICS-3. Darker regions indicate higher density of personas on a logarithmic scale. Colored dots represent selected extreme personas: (a) shows personas selected from all four corners of the political compass, (b) shows personas selected from the economic extremes.\vspace{-4mm}}
    \label{fig:political-compass}
\end{figure}

\section{Introduction}

Large language models (LLMs) have demonstrated remarkable capabilities across various tasks, but their potential political biases remain a significant concern for AI ethics and deployment. This is particularly crucial for content moderation systems, where biased models could disproportionately impact detection rates across different user groups, compromising the overall fairness of content moderation systems.
While pretraining models on politically diverse corpora has been shown to improve downstream task fairness \cite{feng-etal-2023-pretraining}, this approach requires significant computational resources that are often inaccessible to many researchers and organizations. 

Recent work has explored persona-based prompting as a more accessible alternative for introducing diversity in model outputs. \citet{frohling2024personas} showed that injecting synthetic personas into prompts can increase data annotation diversity, while \citet{bernardelle2024mappinginfluencingpoliticalideology} showed that this approach can be used to influence the original political viewpoint of LLMs. However, less attention has been paid to understanding how these political perspectives manifest in practical applications.

Our work bridges this gap by investigating how persona-based prompting can influence real-world applications of LLMs in multimodal hate speech detection, specifically focusing on hateful memes. By having a model assume the perspective of different personas mapped to various points on the Political Compass Test (PCT) \cite{feng-etal-2023-pretraining, rottger-etal-2024-political}, we examine whether their political positioning influences classification decisions for hate speech in memes. This methodology allows us to investigate whether this approach can achieve similar benefits to those observed in previous political pretraining approaches while being more computationally efficient and accessible.
Our investigation addresses two questions:

\begin{itemize}
    \item[\textbf{RQ1:}] How do personas' positions on the PCT correlate with their classification decisions in a hate speech detection task?
    \item[\textbf{RQ2:}] Does injecting explicit political leaning into persona descriptors alter their classification behavior?
\end{itemize}

We observe that there is no correlation between persona's positions on the political compass and their decisions when classifying potentially harmful content. This held true across our experiments, even after strengthening the ideological components of the personas through explicit descriptors. Furthermore, while our investigation confirms that LLMs can display distinct political leanings when directly questioned about political topics, our results suggest that, when using persona-based prompting, these inherent biases play a small role in classification scenarios. However, since our study focused specifically on hate speech detection across a limited set of datasets, these findings may not generalize to other tasks where political biases could have stronger effects.

\section{Related Work}
\paragraph{\textbf{LLM bias and downstream task fairness}}
Previous work has shown that political biases in LLMs can significantly impact fairness in downstream tasks. \citet{feng-etal-2023-pretraining} showed that pretraining data's political orientation affects model performance in tasks like hate speech detection and misinformation classification, with models exhibiting different biases based on their training corpora. Their work revealed that models pretrained on politically diverse sources show varying patterns in how they treat different identity groups and partisan content. This suggests that LLMs encode and propagate political biases from their training data, which can lead to unfair treatment of certain groups or perspectives in practical applications. 
While pretraining on politically diverse data can help address these biases, such approaches often require substantial data and computational resources.

\paragraph{\textbf{Persona-based prompting}}
Recent work has explored persona-based prompting as a more accessible alternative for controlling model behavior and mitigating bias. \citet{frohling2024personas} introduced this approach to increase diversity in annotation tasks by injecting different persona descriptions into model prompts, showing that persona descriptions can help elicit a wider range of valid perspectives. Building on this, \citet{bernardelle2024mappinginfluencingpoliticalideology} showed that persona-based prompting can effectively modulate models' political orientations without requiring expensive retraining. Their work revealed that carefully crafted persona prompts can shift model responses along the political spectrum, offering a promising method for achieving political diversity in language model outputs. This prompting-based approach provides a more flexible and resource-efficient way to control model behavior than pretraining-based methods, while still maintaining the ability to capture diverse political perspectives. While this body of work establishes the effectiveness of persona-based prompting, our work extends this research by evaluating how persona-based political perspectives influence performance on multimodal hateful content detection tasks.

\section{Methodology}
\paragraph{\textbf{Data}}
Our study utilizes three datasets: PersonaHub for synthetic personas, and both the Hateful Memes dataset and MMHS150K for evaluating persona-based moderation capabilities.

We employ the complete collection of 200,000 synthetic personas from PersonaHub \cite{ge2024scaling}, which is publicly available.\footnote{\url{https://huggingface.co/datasets/proj-persona/PersonaHub}}
We map these personas onto the PCT to understand their ideological distribution along social and economic axes, by assessing their political perspectives through 62 distinct statements across six domains. 
The Hateful Memes dataset \cite{kiela2020hateful} consists of 10,000 multimodal memes labeled as either hateful or non-hateful, with additional fine-grained annotations\footnote{\url{https://github.com/facebookresearch/fine_grained_hateful_memes}} for protected groups and attack types provided separately by the dataset authors. Finally, MMHS150K \cite{gomez2020exploring} provides 150,000 manually annotated tweets across six categories of possible attacks.

\paragraph{\textbf{Vision model}}
For our multimodal analysis, we utilize IDEFICS-3 \cite{laurenccon2024building}, a Vision-Language Model (VLM) built on Meta's Llama-3.1-8B-Instruct architecture. 
This choice is motivated by its strong zero-shot performance on multimodal tasks and relatively small parameter count. Furthermore, its architecture, being based on Llama 3.1, allows for better consistency when comparing results with text-only experiments using similar model families.

\paragraph{\textbf{Experimental setup}}
Our experimental design consists of two studies.
In Study 1, from the 200,000 personas we selected 60 (15 per quadrant) based on a scoring system that balanced positional extremity with ideological consistency (see Appendix \ref{apdx:persona_selection}), as shown in Figure \ref{fig:diagonal}. Using these personas, we conducted hateful content classification of 1,000 randomly selected memes from each dataset (Hateful Memes and MMHS150K) using standardized prompt templates (detailed in Appendix \ref{apdx:persona_prompt_templates}). For each persona, we collected binary harm classification, targeted group identification, and attack method categorization.

To control for potential training data contamination, we replicated the experiment on the Hateful Memes test set, whose labels, to the best of our knowledge, are not publicly available, unlike those of MMHS150K.

In Study 2, we focused solely on economic left-right positioning by selecting a sample of 40 personas drawn from the economic extremes (20 each from leftmost and rightmost positions). We then amplified their political orientations by explicitly labeling each persona with its respective ideological leaning (i.e., making right-wing personas more explicitly right-wing and left-wing personas more explicitly left-wing).
The choice of directly testing personas with explicitly labeled ideological leanings was motivated by Study 1's revelation of minimal correlation between political positioning and classification decisions. Study 2 thus served as a test of this finding, examining whether deliberately amplified personas could override the observed independence between political orientation and classification behavior.

This design allows us to evaluate both the baseline impact of political positioning on hate speech detection and the effect of explicit ideological manipulation while reducing the complexity of the annotation process by removing the social dimension from consideration.

\paragraph{\textbf{Agreement analysis}}
To quantify classification consistency across political orientations, we calculated pairwise Cohen's Kappa scores between all personas, resulting in 1,770 unique agreement measurements for Study 1's 60 personas. We analyzed these measurements at two levels: \textit{intra-quadrant agreement} (average agreement between personas within the same political quadrant) and \textit{inter-quadrant agreement} (average agreement between personas from different quadrants). This approach enables us to determine whether personas with similar political orientations exhibit more consistent classification behavior compared to those with opposing viewpoints. For Study 2, we applied the same agreement analysis methodology but focused on comparing agreement patterns between and within just two groups (economic left versus right) to examine whether explicit ideological amplification affects classification consistency.

\paragraph{\textbf{Computational resources}}
All experiments were conducted on a single NVIDIA RTX 4090 GPU. PCT answer generation for 200,000 personas required approximately 48 hours, while meme classification with 60 and 40 personas took 8 and 6 hours respectively. Total computation time across all experiments was approximately 80 hours.

\begin{table}[t]
\centering
\renewcommand{\arraystretch}{1.2}
\caption{Classification performance of IDEFICS-3 using persona-based prompting on the Hateful Memes and MMHS150K datasets. Harmfulness detection is a binary classification task (True/False), while Target Group identification and Attack Method detection are multi-class classification tasks.}
\begin{tabular}{lccc}
\toprule
\textbf{Category} & \textbf{Accuracy} & \textbf{Macro F1} & \textbf{Weighted F1} \\
\midrule
\midrule
\multicolumn{4}{c}{\textbf{Hateful Memes}} \\
\cmidrule(lr){1-4}
Harmfulness (T/F) & 0.908 & 0.890 & 0.907 \\
Target Group & 0.876 & 0.733 & 0.874 \\
Attack Method & 0.743 & 0.310 & 0.724 \\
\midrule
\multicolumn{4}{c}{\textbf{MMHS150K}} \\
\cmidrule(lr){1-4}
Harmfulness (T/F) &  0.593 & 0.555 & 0.623 \\
Target Group & 0.540 & 0.296 & 0.587 \\
\bottomrule
\end{tabular}
\label{tab:visual-performance}
\vspace{-0.2mm}
\end{table}

\section{Results}
\paragraph{\textbf{Vision model performance}}
Before analyzing agreement patterns between different personas, we first evaluated IDEFICS-3's hate speech detection capabilities when employing our persona-based prompting approach. This initial performance assessment is crucial, as meaningful analysis of persona agreement patterns requires the model to demonstrate competent classification abilities. 
To assess whether the introduction of personas impacts model performance, we also ran a baseline evaluation using standard classification prompts without personas. Both approaches achieved nearly identical performance (detailed in Appendix \ref{apdx:baseline_performance}), indicating that persona-based prompting preserves the model's classification capabilities.
As shown in Table \ref{tab:visual-performance}, IDEFICS-3 prompted with persona descriptions obtains strong performance on the Hateful Memes dataset, achieving high scores in harmfulness detection (accuracy: 0.908, F1: 0.890) and target identification (accuracy: 0.876, F1: 0.733), with performance gradually declining for more nuanced tasks like attack method classification (accuracy: 0.743, macro F1: 0.310). 
The model shows moderate performance on the MMHS150K dataset with accuracy scores around 0.5-0.6 across both tasks. While these results exceed random chance, the substantial drop in performance could stem from MMHS150K's broader scope and more nuanced content, making it a more demanding benchmark for hate speech detection.

These results shows that IDEFICS-3 can effectively perform hate speech detection while maintaining distinct persona-based perspectives, though with varying degrees of success across different tasks and datasets, providing a solid foundation for our main investigation into how these personas might agree or disagree in their classifications.

\begin{figure}[t]
    \centering
    \begin{subfigure}[b]{0.48\linewidth}
        \centering
        \includegraphics[width=\textwidth]{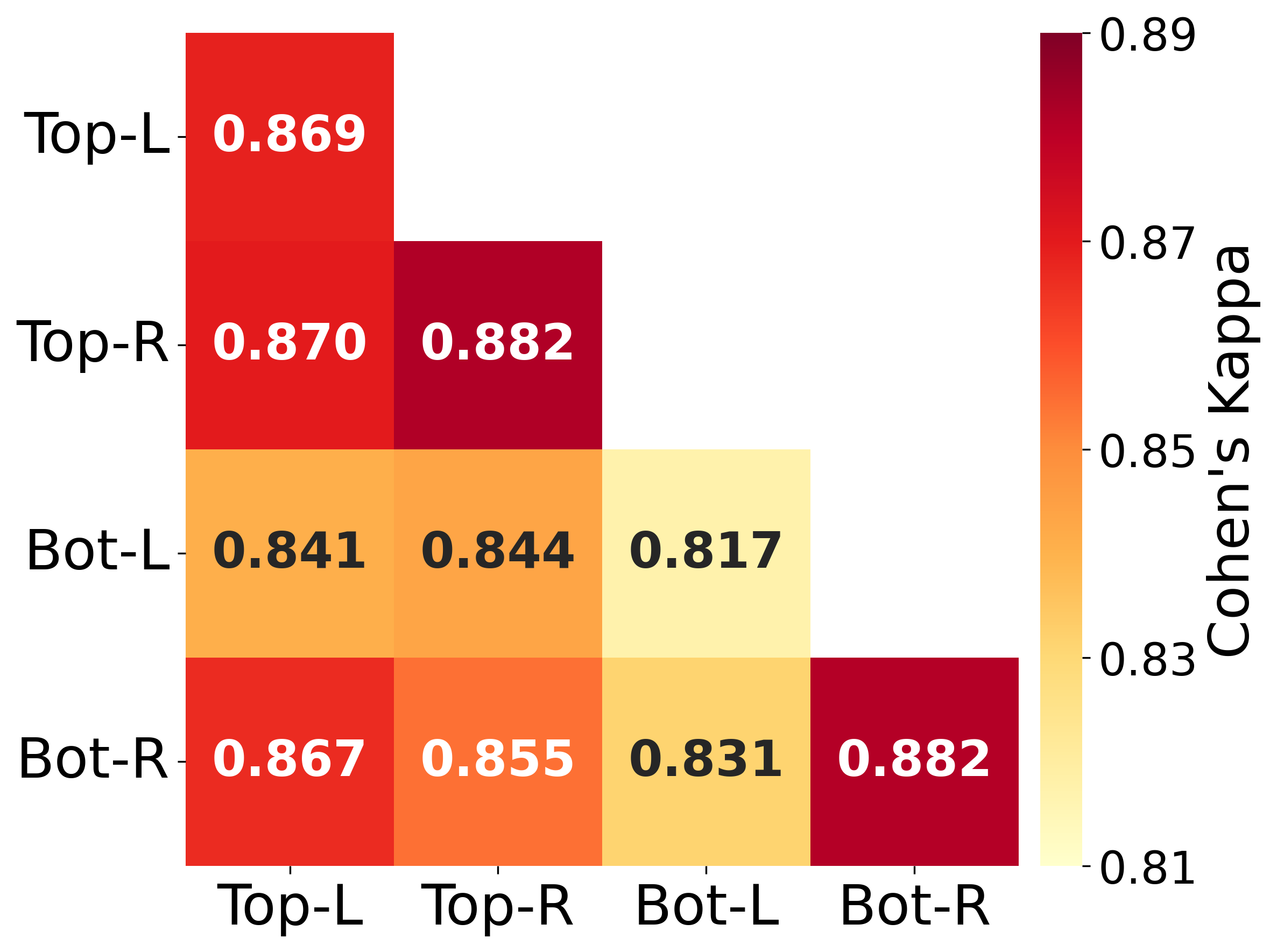}
        \caption{Hateful Memes}
        \label{fig:matrix-a}
    \end{subfigure}
    \hfill
    \begin{subfigure}[b]{0.48\linewidth}
        \centering
        \includegraphics[width=\textwidth]{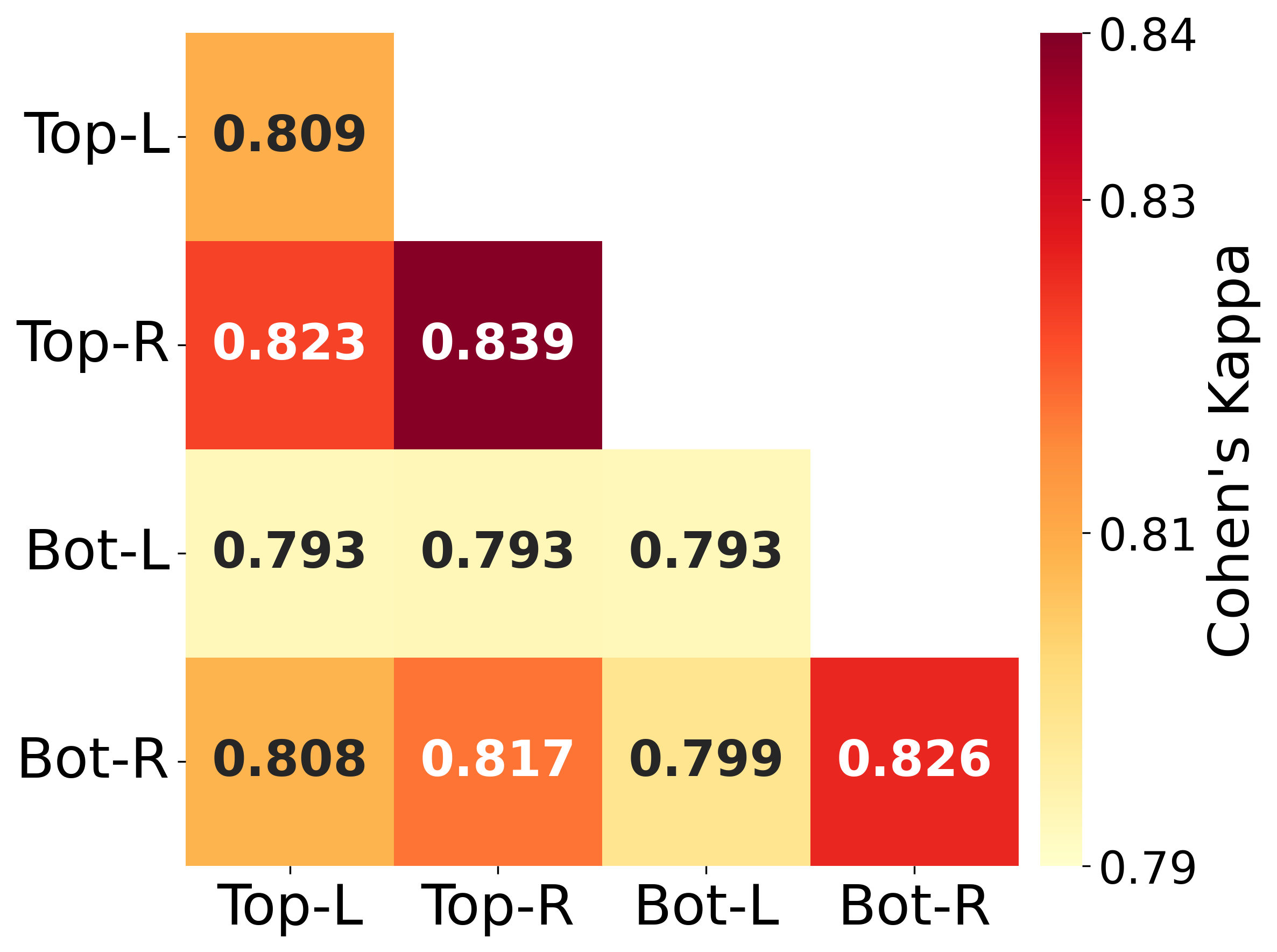}
        \caption{MMHS150K}
        \label{fig:matrix-b}
    \end{subfigure}
    \caption{Matrix showing Cohen's kappa scores for classification agreement between personas from different political quadrants. Diagonal shows intra-quadrant agreement, while off-diagonal elements show inter-quadrant agreement.    \vspace{-4mm}}
    \label{fig:matrix}
\end{figure}

\begin{figure*}[t]
    \centering
    \begin{subfigure}[b]{0.24\linewidth}
        \centering
        \includegraphics[width=\textwidth]{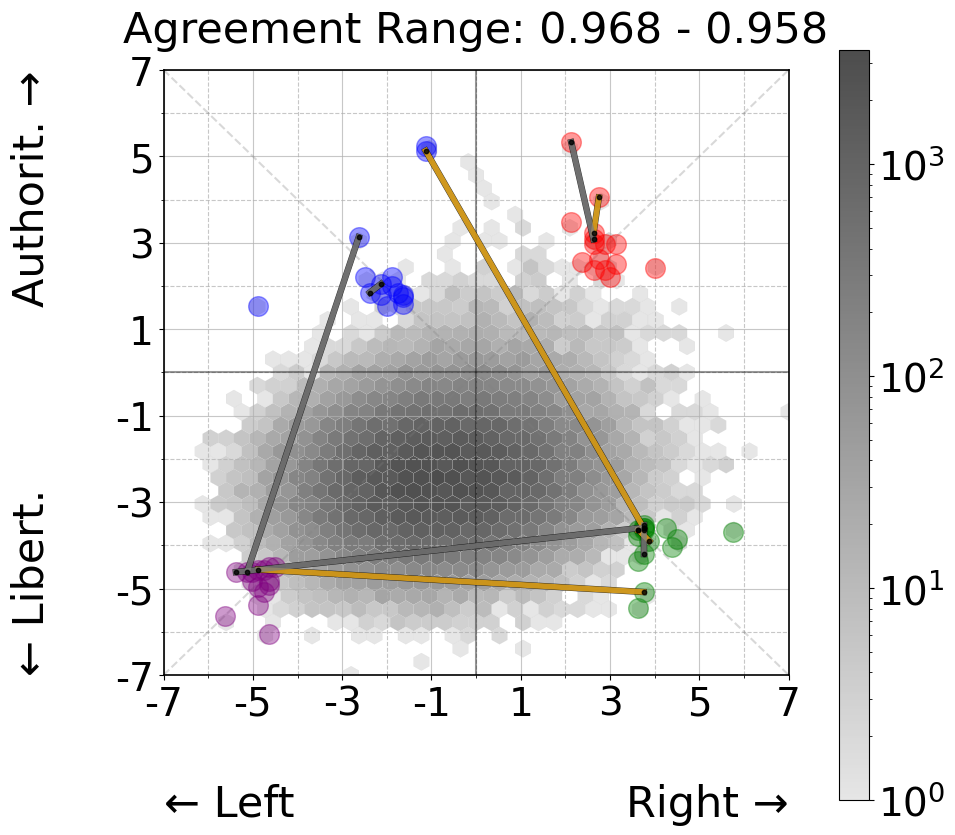}
        \label{fig:subfig1}
    \end{subfigure}
    \hfill
    \begin{subfigure}[b]{0.24\linewidth}
        \centering
        \includegraphics[width=\textwidth]{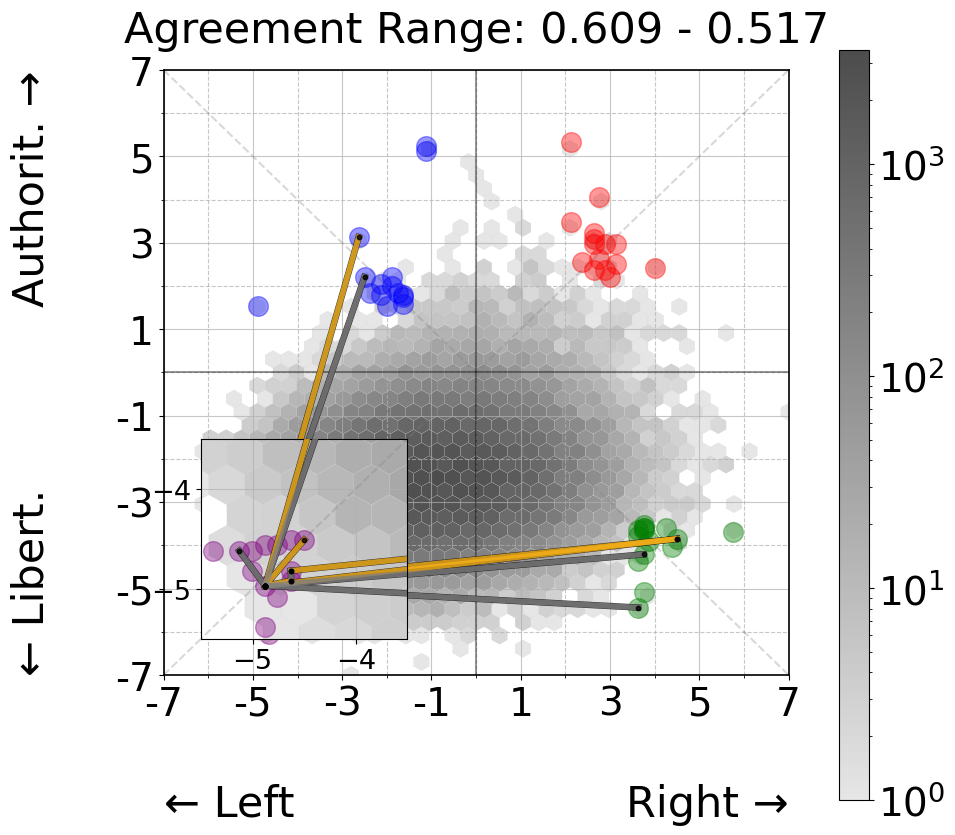}
        \label{fig:subfig2}
    \end{subfigure}
    \hfill
    \begin{subfigure}[b]{0.24\linewidth}
        \centering
        \includegraphics[width=\textwidth]{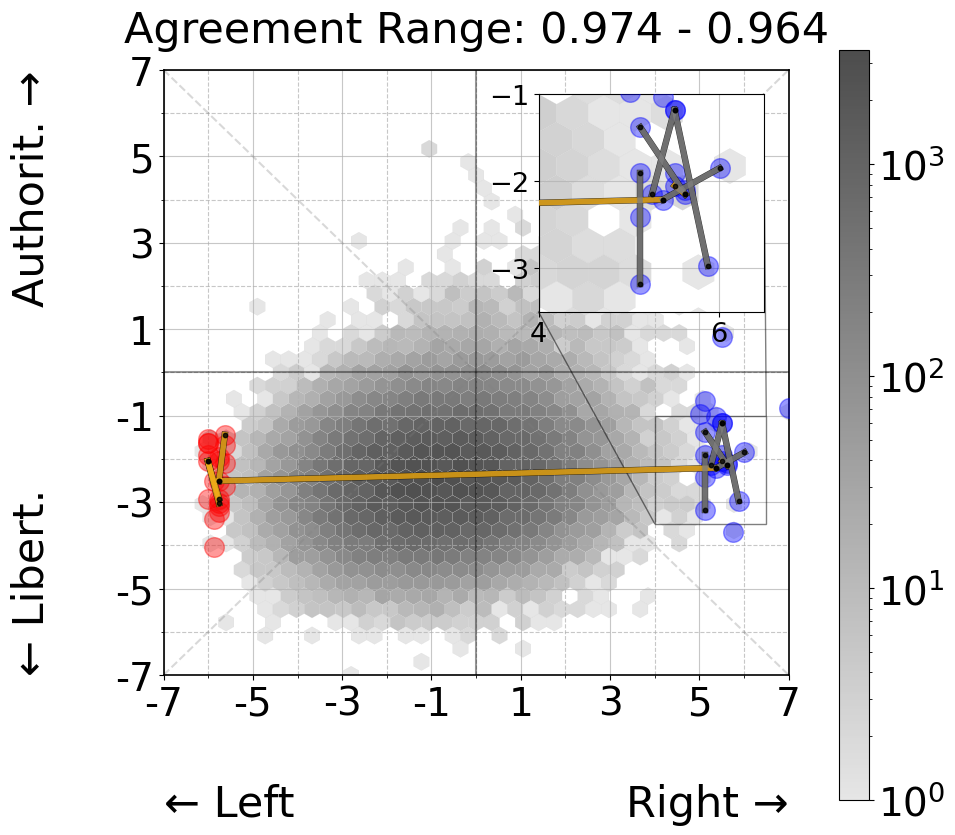}
        \label{fig:subfig3}
    \end{subfigure}
    \hfill
    \begin{subfigure}[b]{0.24\linewidth}
        \centering
        \includegraphics[width=\textwidth]{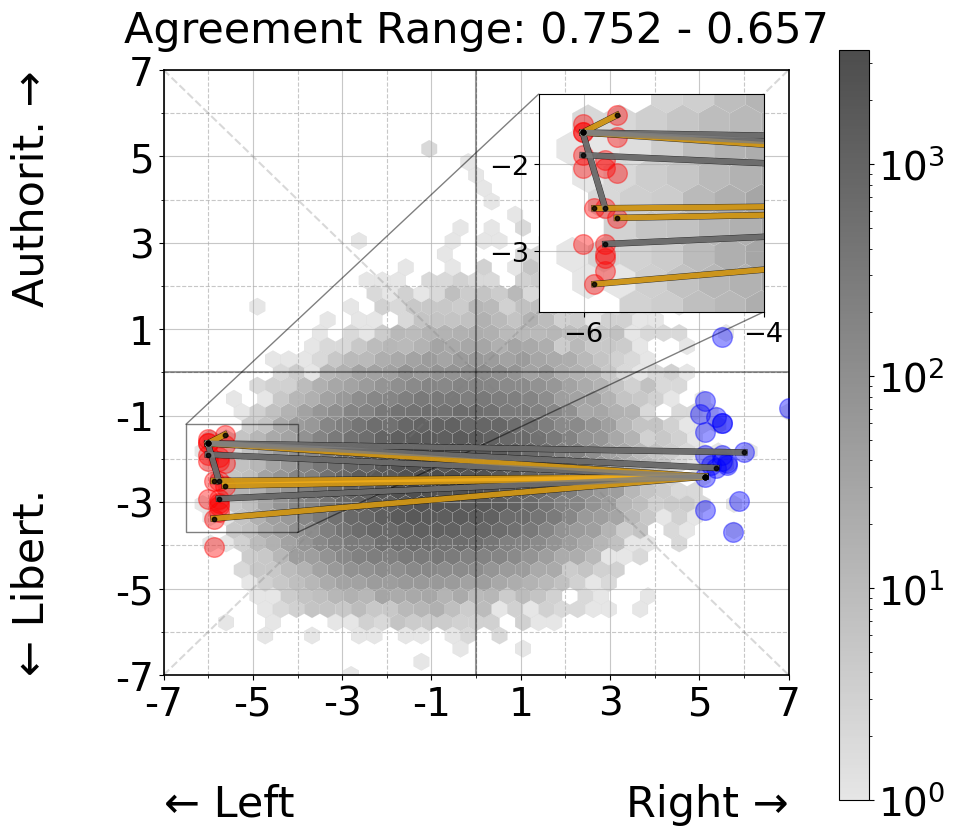}
        \label{fig:subfig4}
    \end{subfigure}
    \vspace{-12pt}
    \caption{Agreement patterns between personas on the Hateful Memes dataset. The left two plots show highest and lowest agreements between the 60 personas across all political quadrants , while the right two plots show the same for the 40 personas from economic extremes. Yellow lines indicate top 5 strongest/weakest agreement pairs, grey lines show the next 5 pairs.}
    \label{fig:agreement-patterns}
\end{figure*}

\paragraph{\textbf{Study 1 - Persona Agreement}}
Our analysis of agreement between the 60 selected personas reveals surprisingly consistent classification behavior across political quadrants. For the Hateful Memes dataset, as shown in Figure \ref{fig:matrix-a}, we observe high agreement both within quadrants (mean Cohen's $k$ = 0.863 ± 0.032 SD) and between quadrants  (mean Cohen's $k$ = 0.851 ± 0.015 SD). The minimal difference between intra- and inter-quadrant agreement suggests that a persona's political positioning has little influence on their classification decisions.
Results from the MMHS150K dataset (Figure \ref{fig:matrix-b}) mirror these patterns, with comparable levels of agreement both within quadrants (diagonal mean $k$ = 0.817, SD = 0.020) and between quadrants (off-diagonal mean $k$ = 0.806, SD = 0.013). The consistency across the two datasets suggests that these agreement patterns might reflect genuine characteristics of persona-based classification rather than dataset-specific artifacts.

Furthermore, when replicating our analysis on the Hateful Memes test set, we found nearly identical agreement patterns (diagonal mean = 0.831, off-diagonal mean = 0.828), indicating that the observed consistency in classification behavior is not due to model memorization of training data.

These findings suggest that persona-based prompting alone may not be sufficient to replicate the ideological influences on hate speech detection observed with political pretraining approaches. Despite clear differences in political positioning, personas demonstrate consistent classification behavior, suggesting that more direct interventions - such as specialized pretraining on politically diverse corpora - may be needed to impact how content is evaluated for harmfulness meaningfully. This raises important questions about the limitations of prompt-based approaches compared to pretraining methods for achieving diverse classifications on hate speech detection tasks. 
While LLMs can display distinct political leanings when directly questioned about political topics, when using persona-based prompting, these inherent biases play a smaller role in classification scenarios than we initially expected.

\paragraph{\textbf{Study 2 - Persona political injection}}
In Study 2, we examined whether explicit ideological amplification of economically left and right personas would increase divergence in classification behavior. 

For the Hateful Memes dataset, agreement levels remained high and uniform (right-leaning k = 0.906, left-leaning k = 0.879, inter-group k = 0.875), closely mirroring Study 1 and suggesting minimal impact of ideological reinforcement on classification.

Analysis of the MMHS150K dataset showed stronger intra-group cohesion compared to inter-group agreement. Right-leaning personas demonstrated the highest internal agreement (k = 0.859), followed by left-leaning personas (k = 0.849). The inter-group agreement was lower (k = 0.807). This distinction might be attributed to MMHS150K's broader scope and more nuanced content, which could provide more opportunities for ideological differences to manifest in classification decisions.

Nevertheless, the overall high agreement scores indicate that persona-induced political ideology, even when explicitly emphasized, plays a relatively minor role in hate speech detection.

\paragraph{\textbf{Agreement Patterns}}
To gain a deeper understanding of how agreement manifests between personas, we analyzed strong and weak agreement patterns across two experimental conditions. For the 60 personas distributed across all political quadrants (Figure \ref{fig:agreement-patterns}, left panels), we observe no clear ideological clustering in agreement patterns. The highest agreement pairs (0.968 - 0.958) span both within and across quadrants, connecting ideologically distant personas just as frequently as ideologically aligned ones. Similarly, the weakest agreement pairs (0.609 - 0.517) show no consistent pattern with respect to political positioning, with low agreement occurring both within and between quadrants. This supports our earlier finding that political quadrant placement has minimal impact on classification decisions.

However, when examining the 40 personas from economic extremes (Figure \ref{fig:agreement-patterns}, right panels), a different pattern emerges. While we still observe a strong agreement between opposing economic positions (yellow line crossing between left and right), the majority of high-agreement pairs (0.974 - 0.964) cluster within their respective economic sides. We observe a similar pattern when examining weaker agreement pairs (0.752 - 0.657), where most instances of disagreement occur between personas from opposing economic positions. This suggests that while explicit economic positioning may not have a strong impact on classification behavior, it does introduce some systematic variation in agreement patterns that wasn't apparent in the more general political quadrant analysis.

\section{Conclusion}
Our research reveals that political ideology, when implemented through persona-based prompting, has minimal impact on hate speech detection in multimodal contexts. The study found consistent classification patterns across different political orientations, even when these orientations were deliberately emphasized. This suggests that the underlying mechanisms governing hate speech detection in LLMs may be deeply embedded in the model's training, and not really susceptible to prompt-based modification. Therefore, persona-based prompting may be insufficient for capturing ideological nuances in content moderation systems.

\paragraph{\textbf{Limitations}}
The scope and generalizability of our findings are primarily constrained by our reliance on the IDEFICS-3 architecture and the two chosen datasets, and may not generalize to other models or different datasets.
Additionally, our persona-based approach might not fully capture the complex decision-making patterns of human content moderators, mitigated by the fact that we are shifting the focus to political leaning instead of the specific persona descriptions. Furthermore, while we attempted to control for potential training data contamination by testing on held-out datasets, we cannot completely rule it out.
Finally, the study relies heavily on Western political frameworks through the PCT, which may not fully capture ideological distinctions relevant to other cultural contexts.

\paragraph{\textbf{Future Work}}
Future research could extend this work in several directions: 1) investigating whether our observations hold across different model architectures and sizes to establish the generalizability of our results; 
2) examining how persona-based prompting influences other content moderation tasks; 3) conducting a direct experimental comparison between persona-based prompting and political pretraining methods to quantitatively assess their relative effectiveness in achieving politically diverse perspectives.

\paragraph{\textbf{Acknowledgments}}
This work is partially supported by the Swiss National Science Foundation (SNSF) under contract number CRSII5\_205975 and by an Australian Research Council (ARC) Future Fellowship Project (Grant No. FT240100022).

\bibliographystyle{ACM-Reference-Format}
\bibliography{personaClassification}

\appendix

\section{Persona Selection Methodology}
\label{apdx:persona_selection}

Our persona selection process aimed to identify individuals with well-defined political positions while maintaining representation across the political spectrum. We developed a systematic scoring approach that favors personas with both extreme and clearly aligned ideological stances within their respective quadrants.

\subsection{Selection Criteria and Metrics}

For each persona $p$ with coordinates $(x, y)$ on the Political Compass Test (PCT), we computed two fundamental metrics:

\subsubsection{Extremity Score}
We quantified the extremity of a persona's political position using their Euclidean distance from the origin:

\begin{equation*}
    E(p) = \sqrt{x^2 + y^2}
    \label{eq:extremity}
\end{equation*}

This metric favors personas with strong political convictions, as indicated by their distance from centrist positions.

\subsubsection{Quadrant Alignment Score}
To ensure selected personas clearly represent their quadrant's ideology, we calculated their alignment with the quadrant's diagonal axis. The alignment score $A(p)$ is computed differently for each quadrant pair:

For top-right (TR) and bottom-left (BL) quadrants:
\begin{equation*}
    A_{\text{TR,BL}}(p) = \frac{|y - x|}{\sqrt{2}}
    \label{eq:align_tr_bl}
\end{equation*}

For top-left (TL) and bottom-right (BR) quadrants:
\begin{equation*}
    A_{\text{TL,BR}}(p) = \frac{|y + x|}{\sqrt{2}}
    \label{eq:align_tl_br}
\end{equation*}

This measure represents the perpendicular distance from the persona's position to their quadrant's principal diagonal, normalized by $\sqrt{2}$ to maintain consistency with the extremity score scale.

\subsection{Composite Selection Score}

We combined these metrics into a final selection score $S(p)$ using a weighted sum with quadrant-specific normalization:

\begin{equation*}
    S(p) = (1-w) \cdot \hat{E}_q(p) + w \cdot (1-\hat{A}_q(p))
    \label{eq:composite}
\end{equation*}

Where:
\begin{itemize}
    \item $\hat{E}_q(p)$ is the extremity score normalized within quadrant $q$
    \item $\hat{A}_q(p)$ is the alignment score normalized within quadrant $q$
    \item $w$ is the diagonal weight parameter (set to 0.4 in our experiments)
\end{itemize}

The normalization is performed separately within each quadrant $q$:

\begin{equation*}
    \hat{E}_q(p) = \frac{E(p)}{\max_{p \in P_q} E(p)}
    \label{eq:norm_extremity}
\end{equation*}

\begin{equation*}
    \hat{A}_q(p) = \frac{A(p)}{\max_{p \in P_q} A(p)}
    \label{eq:norm_alignment}
\end{equation*}

Where $P_q$ represents the set of all personas in quadrant $q$. This quadrant-specific normalization ensures fair comparison of personas within each ideological region.

\subsection{Selection Process}

For each quadrant, we:
\begin{enumerate}
    \item Computed the composite score $S(p)$ for all personas
    \item Ranked personas by their composite scores
    \item Selected the top 15 personas per quadrant
\end{enumerate}

This methodology yielded a balanced set of 60 personas (15 per quadrant) that exhibit both strong ideological positions and clear quadrant alignment, ensuring robust representation across the political spectrum while minimizing ambiguous or centrist positions.

\section{Personas and Prompt Templates}
\label{apdx:persona_prompt_templates}
This appendix details the prompt templates and evaluation methodology used in our experimental studies for classifying harmful content.

\subsection{Prompt Templates}
To perform the classification, we used \textit{Study 1 Prompt Template}, inserting each of the 60 selected persona descriptions into the [PERSONA] placeholder when prompting the visual model\footnote{Note: The provided prompt templates were used for classification on the Hateful Memes dataset. Similar templates were used for MMHS150K with labels modified to match the dataset's annotation schema.}.

\subsubsection{Study 1 Prompt Template}
The following template was used for classification in Study 1, with [PERSONA] replaced by each of the 60 selected persona descriptions:

\begin{quote}
Taking on the perspective of [PERSONA], analyze this meme in two steps:
\begin{enumerate}
    \item First, classify if this meme is harmful or not:\\
          Answer with only: HARMFUL or NOT\_HARMFUL
    \item If you answered HARMFUL:\\
          Targeted groups affected (TG:) (choose 1): disability, race, religion, nationality, sex\\
          Attack methods used (AM:) (choose 1): contempt, mocking, inferiority, slurs, exclusion, dehumanizing, inciting\_violence
\end{enumerate}
When executing this task, think carefully about how a person with the described perspective would interpret the meme.
\end{quote}

\subsubsection{Study 2 Prompt Template}
The following template was used for classification in Study 2, focusing on explicit economic left-right positioning:

\begin{quote}
Analyze this meme taking on the perspective of the following politically [LEFT or RIGHT] leaning persona: [PERSONA].
\begin{enumerate}
    \item First, classify if this meme is harmful or not:\\
          Answer with only: HARMFUL or NOT\_HARMFUL
    \item If you answered HARMFUL:\\
          Targeted groups affected (TG:) (choose 1): disability, race, religion, nationality, sex\\
          Attack methods used (AM:) (choose 1): contempt, mocking, inferiority, slurs, exclusion, dehumanizing, inciting\_violence
\end{enumerate}
When executing this task, think carefully about how a person with the described [LEFT or RIGHT] leaning perspective would interpret the meme.
\end{quote}

\subsection{Example Personas}
To provide concrete examples of the personas used in our studies, we present below a selection of three representative personas from each quadrant of the political compass.

\paragraph{Authoritarian Right (Top-Right) Examples}
\begin{itemize}
    \item A local politician who values the pastor's endorsement for their conservative agenda
    \item A fresh graduate who idolizes her supervisor for his conservative science perspectives
    \item A business developer trying to bring new investments to the region, regardless of environmental cost
\end{itemize}
\paragraph{Authoritarian Left (Top-Left) Examples}
\begin{itemize}
    \item A high-ranking military strategist who believes in a strong defense system
    \item A traditionalist school principal who believes in the old-school methods and resists change
    \item A young adult who is searching for meaning in life and is drawn to the cult leader's philosophy
\end{itemize}
\paragraph{Libertarian Right (Bottom-Right) Examples}
\begin{itemize}
    \item A rival department head who is skeptical about the effectiveness of e-learning
    \item A rival fuel broker vying for the same clients, employing aggressive tactics to win contracts
    \item A representative from a telecommunications company advocating for less restrictive regulations on satellite deployment
\end{itemize}
\paragraph{Libertarian Left (Bottom-Left) Examples}
\begin{itemize}
    \item A graduate student advocating for fair working conditions and organizing protests
    \item A discriminant sports fan who doesn't follow college basketball
    \item A socialist advocate who argues that free trade perpetuates inequality and exploitation
\end{itemize}

\section{Baseline Model Performance without Personas}
\label{apdx:baseline_performance}
\begin{table}[ht]
\centering
\renewcommand{\arraystretch}{1.2}
\caption{Classification performance of IDEFICS-3 WITHOUT using persona-based prompting on the Hateful Memes and MMHS150K datasets. Harmfulness detection is a binary classification task (True/False), while Target Group identification and Attack Method detection are multi-class classification tasks.}
\begin{tabular}{lccc}
\toprule
\textbf{Category} & \textbf{Accuracy} & \textbf{Macro F1} & \textbf{Weighted F1} \\
\midrule
\midrule
\multicolumn{4}{c}{\textbf{Hateful Memes}} \\
\cmidrule(lr){1-4}
Harmfulness (T/F) & 0.919 & 0.898 & 0.916 \\
Target Group & 0.786 & 0.510 & 0.745 \\
Attack Method & 0.725 & 0.190 & 0.650 \\
\midrule
\multicolumn{4}{c}{\textbf{MMHS150K}} \\
\cmidrule(lr){1-4}
Harmfulness (T/F) &  0.607 & 0.581 & 0.621 \\
Target Group & 0.533 & 0.278 & 0.557 \\
\bottomrule
\end{tabular}
\label{tab:baseline-performance}
\end{table}
Table \ref{tab:baseline-performance} shows IDEFICS-3's performance on the classification tasks when using standard prompts without personas. Prompts used are the same as the ones used in Appendix \ref{apdx:persona_prompt_templates} but without the persona-related part.
The model with persona-prompting achieves comparable or better results across all metrics. While harmfulness detection shows similar performance (within 1-2 percentage points difference), persona-prompting notably improves performance on more complex tasks - target group identification accuracy increases by 9 percentage points (from 0.786 to 0.876) and macro F1 score improves by 22.3 percentage points (from 0.510 to 0.733). Similarly, for attack method classification, the persona-prompted version shows stronger performance with a 12 percentage point improvement in macro F1 score (from 0.190 to 0.310) and better accuracy (0.743 vs 0.725). These results suggest that persona-based prompting not only enables investigation of potential ideological influences but can also enhance the model's classification capabilities, particularly for more nuanced multi-class tasks.

\end{document}